\documentclass[10pt,twocolumn,letterpaper]{article}

\usepackage{iccv}
\usepackage{times}
\usepackage{epsfig}
\usepackage{graphicx}
\usepackage{amsmath}
\usepackage{amssymb}

\usepackage{pifont}
\newcommand{\cmark}{\ding{51}}
\newcommand{\xmark}{\ding{55}}
\usepackage[symbol]{footmisc}

\usepackage{adjustbox}
\usepackage{array}
\usepackage{tabularx}

\usepackage{booktabs}
\usepackage{multirow}
\usepackage[accsupp]{axessibility}
\usepackage[pagebackref=true,breaklinks=true,letterpaper=true,colorlinks,bookmarks=false]{hyperref}

\iccvfinalcopy % *** Uncomment this line for the final submission

\def\iccvPaperID{10533} % *** Enter the ICCV Paper ID here

\ificcvfinal\fi

\begin{document}

%%%%%%%%% TITLE
\title{Skip-Plan: Procedure Planning in Instructional Videos \\ via Condensed Action Space Learning}

\author{
    Zhiheng Li\textsuperscript{1},
    Wenjia Geng\textsuperscript{2},
    Muheng Li\textsuperscript{1},    
    Lei Chen\textsuperscript{3},
    Yansong Tang\textsuperscript{2}\thanks{\;indicates the corresponding author.}, 
    Jiwen Lu\textsuperscript{1,4},
    Jie Zhou\textsuperscript{1,4}\\
    \textsuperscript{1} Department of Automation, Tsinghua University\\
    \textsuperscript{2} Shenzhen International Graduate School, Tsinghua University\\
    \textsuperscript{3} Beijing University of Science and Technology\\
    \textsuperscript{4} Beijing National Research Center for Information Science and Technology\\
    {\tt \small \{lizhihan21@,gengwj22@,li-mh20@\}mails.tsinghua.edu.cn,chenlei2022@ustb.edu.cn,}\\
    {\tt \small \{tang.yansong@sz.,lujiwen@,jzhou@\}tsinghua.edu.cn}\\
}

% \author{Zhiheng Li\\
% Tsinghua University\\
% {\tt\small lizhihan21@mails.tsinghua.edu.cn}
% \and

\maketitle
\ificcvfinal\thispagestyle{empty}\fi

%%%%%%%%% ABSTRACT
\begin{abstract}
     In this paper, we propose Skip-Plan, a condensed action space learning method for procedure planning in instructional videos. Current procedure planning methods all stick to the state-action pair prediction at every timestep and generate actions adjacently. Although it coincides with human intuition, such a methodology consistently struggles with high-dimensional state supervision and error accumulation on action sequences. In this work, we abstract the procedure planning problem as a mathematical chain model. By skipping uncertain nodes and edges in action chains, we transfer long and complex sequence functions into short but reliable ones in two ways. First, we skip all the intermediate state supervision and only focus on action predictions. Second, we decompose relatively long chains into multiple short sub-chains by skipping unreliable intermediate actions. By this means, our model explores all sorts of reliable sub-relations within an action sequence in the condensed action space. Extensive experiments show Skip-Plan achieves state-of-the-art performance on the CrossTask and COIN benchmarks for procedure planning. 
\end{abstract}

%%%%%%%%% BODY TEXT
\section{Introduction}
\label{sec:intro}
\begin{figure}[t]
  \centering
  \includegraphics[width=\linewidth]{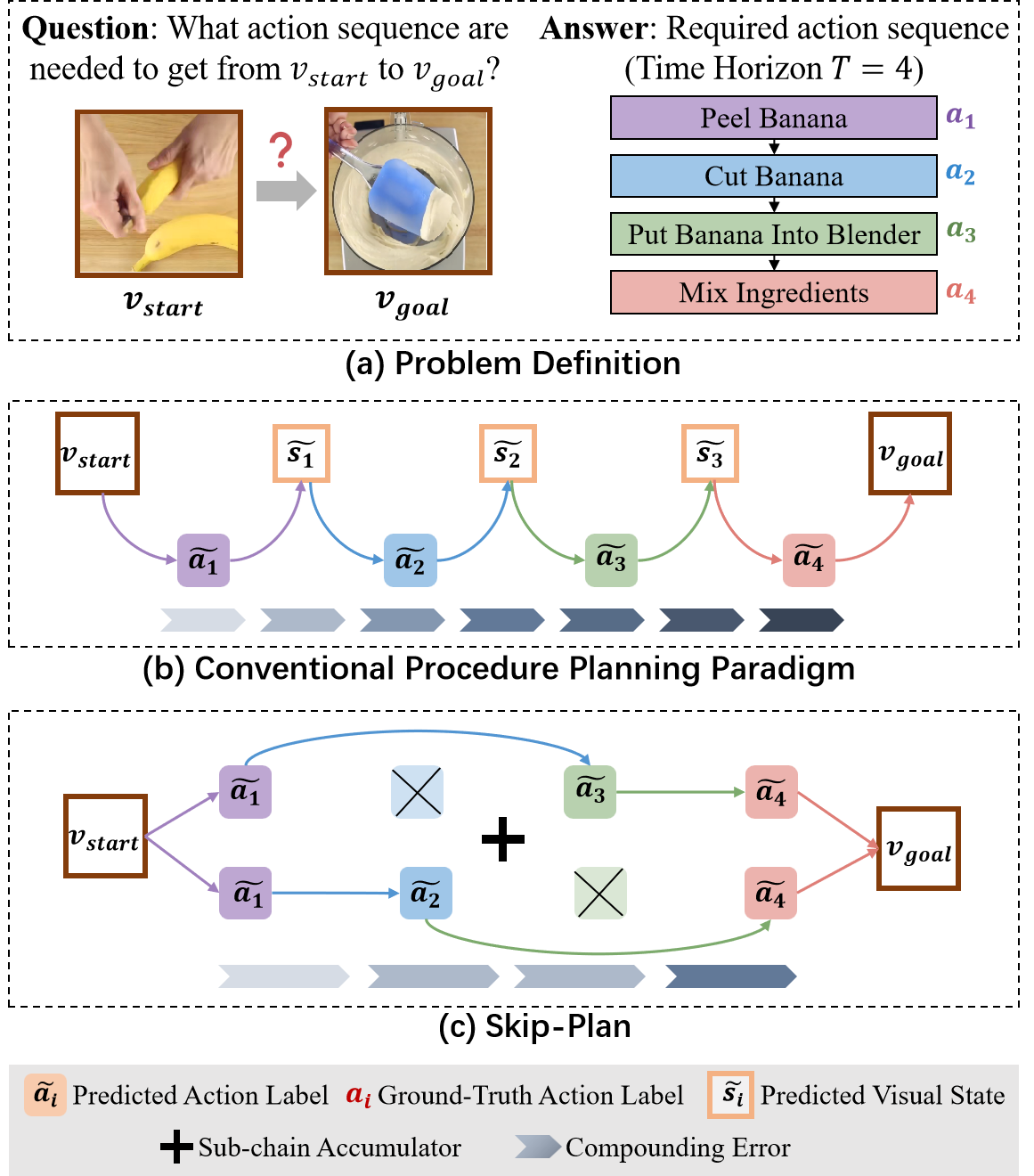}
  \caption{\textbf{Comparisons of the conventional paradigm and our proposed Skip-Plan in procedure planning. } (a) Problem definition: given the initial and goal visual observations ($v_{start}$ and $v_{goal}$), procedure planning is to answer an action sequence to reach the indicated goal with a pre-defined time horizon $T$. (b) Conventional paradigm predicts a state-action pair $\{\Tilde{s}_t,\Tilde{a}_t\}$ at every timestep $t$, and repeat it step-by-step. (c) Our Skip-Plan skips state supervision and extracts multiple sub-chains by skipping most intermediate actions. The final action sequence is accumulated from all predicted sub-chains. Skip-Plan significantly reduces the overall compound error, indicated by the shallower darkness of the chevron arrows. }
  \vspace{-1pt}
  \label{fig:1}
\end{figure}

Given the initial and goal visual observations which are curated from instructional videos \cite{crosstask, coin}, the task of procedure planning is to predict a sequence of actions to reach the indicated goal \cite{procedure_planning1}, as illustrated in Figure \ref{fig:1}(a). This is an emerging research direction in the field of video understanding. Procedure planning has worldwide applications and assists humans with instruction-following tasks, such as cooking, assembling furniture, electronics repair, etc. 

% Procedure planning is the task where we predict a sequence of actions to reach the indicated goal based on the initial and goal visual observations \cite{procedure_planning1}, as illustrated in Figure \ref{fig:1}(a). This is an emerging research direction in the field of action planning. Procedure planning is highly useful in assisting humans with instruction-following tasks, such as cooking, assembling furniture, electronics repair, etc. In this work, we study the procedure planning problem in instructional videos, including CrossTask and COIN \cite{crosstask, coin}. 

In the past few years, various approaches have been proposed to solve the procedure planning problem. Most existing methods utilize an autoregressive model with visual state supervision to generate actions one by one \cite{procedure_planning1, procedure_planning2, plate}. Another type of work \cite{p3iv} solves it in a non-autoregressive way with language state supervision. The difficulties of procedure planning have been widely discussed in the previous works. For instance, the sequential orders in procedure planning can be altered but still sensible (\eg, switching the order of ‘pour water’ and ‘pour alcohol’ does not matter in many cases), and the visual states are unstructured and high-dimensional. Thus, we can conclude the following limitations from the existing works. (a) \textbf{State supervision}: the major difficulty in procedure planning is state supervision. Visual images are high-dimensional and complex, with extra misleading information such as background, illumination, viewing point, etc. (b) \textbf{Step-by-step predictions}: the conventional paradigm in procedure planning predicts one state-action pair at every timestep, and repeats it step-by-step, as illustrated in Figure \ref{fig:1}(b). Even though the proposed non-autoregressive model generates whole action sequences in batches \cite{p3iv}, it can be still regarded as a step-by-step method because the model neglects the links between non-adjacent actions. From Figure \ref{fig:1}(b), we can see while predicting an action sequence with a time horizon of 4, the conventional model is equivalently predicting a sequence with a length of 7 (4 action nodes, 3 state nodes). More frequent predictions tend to accumulate a larger compound error along the chain, which lowers the prediction accuracy. 

% In the past few years, various approaches have been proposed to solve the procedure planning problem. Most existing methods utilize an autoregressive model with visual state supervision to generate actions one by one \cite{procedure_planning1, procedure_planning2, plate}. Another type of work \cite{p3iv} solves it in a non-autoregressive way with language state supervision. The difficulties of procedure planning have been widely discussed in the previous works. For instance, the sequential orders in procedure planning can be altered but still sensible (\eg, switching the order of ‘pour water’ and ‘pour alcohol’ does not matter in many cases), and the visual states are unstructured and high-dimensional. Thus, we can conclude the following points from the existing works: (a) the major difficulty in procedure planning is state prediction. Visual images are high-dimensional and complex, with extra misleading information such as background; (b) the conventional paradigm in procedure planning predicts one state-action pair at every timestep, and repeats it step-by-step, as illustrated in Figure \ref{fig:1}(b). Even though the proposed non-autoregressive model generates whole action sequences in batches \cite{p3iv}, it can be still regarded as a step-by-step method because the model neglects the links between non-adjacent actions. 

To overcome these limitations, we propose Skip-Plan, a condensed action space learning method that solves the procedure planning problem through the lens of mathematics, aiming to reduce compound error along action chains. As we have discussed above, existing methods struggle with high-dimensional and complex state supervision. Thus, the compound error along the action chain is increased due to large errors in state predictions. This motivates us to skip all the state nodes in our Skip-Plan framework and only model sequence relations in the condensed action space, as demonstrated in Figure \ref{fig:1}(c). Without the presence of state supervision, a lower compound error results in better performance in Skip-Plan. Although the dependencies between high-level actions without state supervision are very abstract, the algorithm of our Skip-Plan is highly mathematically sensible in the Markov chain model. 

% Distinct from the previous works, we propose Skip-Plan which solves the procedure planning problem through the lens of mathematics, aiming to reduce the compound error along action chains. From Figure \ref{fig:1}(b), we can see the conventional model is equivalently predicting a sequence with a length of 7 (4 action nodes, 3 state nodes) while predicting an action sequence with T = 4. More frequent predictions must accumulate a larger compound error along the chain, which lowers the prediction accuracy. Also, it is well-acknowledged that visual states are much harder to be predicted than actions. Therefore, our Skip-Plan skips all the state nodes and only model sequence relations in the condensed action space, as demonstrated in Figure \ref{fig:1}(c). Although the dependencies between high-level actions without state supervision are very abstract, the algorithm of our Skip-Plan is highly mathematically sensible in the Markov chain model. 

Another convention in procedure planning is we should learn sequential relations between adjacent actions. The Markov chain model inspires us to decouple relatively long chains into multiple short sub-chains, because the compound error can be reduced with shorter chain lengths. The problem here is how to decouple these long chains, because there are various choices of sub-chains with different lengths and we should always choose the most reliable ones under any condition. Given the initial and goal frames as concrete inputs, it is apparent the reliabilities of the initial and final actions are the highest, and the intermediate actions are more likely to be predicted wrong. Therefore, our Skip-Plan decomposes long chains by `skipping' some unreliable intermediate actions, and each sub-chain in our model is composed of the initial action, one unique intermediate action, and the final action. One example for a chain length of 4 is shown in Figure \ref{fig:1}(c). In this way, the reliable sub-relations between non-adjacent actions can be well explored by the network. To model the discriminative relations of all sub-chains, we propose sub-chain decoders, where each decoder only learns one sub-chain by the specific loss design. The final action sequence is aggregated from all sub-chain predictions by a sub-chain accumulator. 

% Another convention in procedure planning is we should learn sequential relations between adjacent actions. The Markov chain model inspires us to decouple relatively long chains into multiple short ones, because the compound error can be reduced with shorter chain lengths. In this way, the reliable sub-relations between non-adjacent actions can be well explored by the network. The problem here is how to decouple these long chains, because there are various choices of sub-chains with different lengths and we should always choose the most reliable ones under any condition. Given the initial and goal frames as concrete inputs, it is apparent the reliabilities of the initial and final actions are the highest, and the intermediate actions are more likely to be predicted wrong. Consequently, each sub-chain in our model is composed of the initial action, one unique intermediate action, and the final action. One example for a chain length of 4 is shown in Figure \ref{fig:1}(c). To model the discriminative relations of all sub-chains, we propose sub-chain decoders, where each decoder only learns one sub-chain by the specific loss design. The final action sequence is aggregated from all the sub-chain predictions by a sub-chain accumulator. 

In summary, our contributions are threefold. (a) Unlike the previous methods which put sophisticated designs on state supervision, Skip-Plan skips unreliable state predictions to boost the model performance in procedure planning. (b) In contrast to the existing methods which only focus on the links between adjacent actions, we decompose long chains into multiple reliable sub-chains by skipping some intermediate actions, and propose sub-chain decoders to learn their discriminative relations. (c) Extensive experiments show Skip-Plan achieves state-of-the-art performance on the CrossTask \cite{crosstask} and COIN \cite{coin} benchmarks for procedure planning.

%-------------------------------------------------------------------------

\section{Related Work}
\label{sec:relatedwork}

\noindent \textbf{Procedure Planning.} Procedure planning in instructional videos was first introduced by Chang \etal \cite{procedure_planning1}. Recently, various methods have been conducted to solve this task. Chang \etal \cite{procedure_planning1} propose a dual dynamic network to conjugate the relationships between states and actions. Sun \etal \cite{plate} utilize a GPT network to generate actions and states one by one. Bi \etal \cite{procedure_planning2} use the model-based reinforcement learning with time-invariant context information to predict state-action pairs autoregressively. Different from the previous autoregressive methods with visual state supervision, Zhao \etal \cite{p3iv} use a non-autoregressive transformer decoder to generate action sequences with language state supervision. Unlike previous works, Skip-Plan generates actions non-adjacently without any state supervision.  

% \noindent \textbf{Procedure Planning.} Procedure planning in instructional videos was first introduced by Chang \etal \cite{procedure_planning1}. Recently, various methods have been conducted to solve this task. Chang \etal \cite{procedure_planning1} propose a dual dynamic network to conjugate the relationships between states and actions. Sun \etal \cite{plate} utilize a GPT network to generate actions and states one by one. Bi \etal \cite{procedure_planning2} use the model-based reinforcement learning with time-invariant context information to predict state-action pairs autoregressively. Different from the previous autoregressive methods with visual state supervision, Zhao \etal \cite{p3iv} use a non-autoregressive transformer decoder to generate action sequences with language state supervision. Unlike any work in procedure planning, Skip-Plan generates actions non-adjacently without any state supervision.  

\noindent \textbf{Instructional Videos Analysis.} Instructional videos are used for learning human behaviours because they are visual representations of human actions. Based on instructional videos, various action-related tasks have been proposed, such as action segmentation\cite{actionseg1, actionseg2,actionseg3,actionseg4,actionseg5,actionseg6}, step localization\cite{steploc1,steploc2,steploc3,steploc4,steploc5}, procedure segmentation\cite{procedureseg1,procedureseg2,procedureseg3,procedureseg4}, video summarization/captioning\cite{videosum1,videosum2,videosum3,videosum4,videosum5,videosum6}, etc. Different from these action tasks in instructional videos, we investigate procedure planning, a goal-conditioned problem that plans a sequence of actions given the initial and goal visual observations. 

\noindent \textbf{Predictive Uncertainty on Sequence Relations. } All types of questions that we are interested in can be modeled as a sequence model, such as images \cite{swin} and languages \cite{bert}. Currently, Transformer \cite{transformer} is the most effective network to model sequence relations, and our model is based on the transformer decoder with a learnable query input\cite{tqn, detr}. However, sequence relations are always distorted by predictive uncertainty\cite{uncertaintysurvey, uncertainty2, uncertainty3, uncertainty4, uncertainty5}, and the sources of predictive uncertainty can be various, including model uncertainty, data uncertainty, etc\cite{uncertainty6, uncertainty7, uncertainty8, uncertainty9, uncertainty10}. In this work, we pay particular attention to data uncertainty, in which the uncertainty directly stems from the data. Data uncertainty is caused by information loss when we measure a data distribution in a space with incorrect dimensions (\eg, we must lose depth information if we measure a three-dimensional object in a two-dimensional space). Therefore, we often avoid high-dimensional data for lower data uncertainty. 

% We will explain why we discard state supervision from the perspective of predictive uncertainty in Section \ref{subsec:nostate}.

\section{Methods}
\label{sec:methods}
\subsection{Problem Formulation}
\label{subsec:problemformulation}
We follow the definition of procedure planning proposed by Chang \etal \cite{procedure_planning1}. Here, we denote the action and the state at timestep $t$ as $a_t$ and $s_t$ respectively. Given the initial visual observation $v_{start}$ and the goal visual observation $v_{goal}$, the model generates a sequence of actions $a_{1:T}$ to reach the indicated goal. In this work, we introduce a novel formulation for the procedure planning task:
\begin{equation}
% \begin{split}
  p(a_{1:T}|v_{s}, v_{g}) = \prod_{t=2}^{T-1} p(a_{t}|a_{1},a_{T})p(a_{1},a_{T}|v_{s},v_{g}),
  \label{eq:1}
% \end{split}
\end{equation}
\noindent where $v_s$ and $v_g$ short for `$v_{start}$' and `$v_{goal}$'. We denote $p(a_{1:T}|v_{start}, v_{goal})$ as the conditional probability distribution of the action sequence given the initial visual observation $v_{start}$ and the goal visual observation $v_{goal}$, $p(a_t|a_1,a_T)$ as the conditional probability distribution of intermediate actions ($2\leq t\leq T-1$) given the initial action $a_1$ and the final action $a_T$, and $p(a_{1},a_{T}|v_{start},v_{goal})$ as the conditional probability distribution of the initial action $a_1$ and the final action $a_T$ given the initial visual observation $v_{start}$ and the goal visual observation $v_{goal}$. Equation \ref{eq:1} reveals the uniqueness of our Skip-Plan method: First, we do not rely on any intermediate state supervision $s_t$ for action generations. Second, every intermediate action prediction $a_t$ $(2\leq t\leq T-1)$ only depends on the initial action $a_1$ and the final action $a_T$. Thus, Skip-Plan does not focus on sequential relations between adjacent actions but predicts action sequences non-adjacently without state supervision. 

\subsection{Action Predictions without State Supervision}
\label{subsec:nostate}
Information does not always help in the decision-making process. Sometimes ambiguous information can even lead to sub-optimal results, and plausible plans can be only made based on reliable information. In procedure planning, two types of supervision can be utilized, including actions $a_t$ and visual states $s_t$. Even though they can be directly generated from each other, the intrinsic properties of these two supervisions are largely different. We evaluate them from the perspectives of the value of information and predictive uncertainty. Visual state supervision is well known to be high-dimensional. In addition to the factor of action class, various factors such as background, viewing point, illumination, etc. can be involved in visual states. Thus, useful information in visual state supervision for action predictions is overwhelmed by many irrelevant factors. In contrast, action labels are the most condensed supervision, without being interrupted by any factors. From this point, we argue the value of information offered by state supervision is extremely low because the portion of information, which is relevant to action predictions, is too small. On the other hand, when models generate high-dimensional predictions such as visual images, large predictive errors are more likely to occur. Thus, the predictive uncertainty of intermediate visual states is significantly higher than the one of condensed action predictions. Combining the effects of the value of information and the predictive uncertainty, we argue the existence of visual state supervision hurts the predictive accuracy of sequential actions. Even though we can use the weak language state supervision to reduce some irrelevant dimensions, it is still too high-dimensional compared to condensed action labels. Therefore, we `skip' intermediate state supervision in our Skip-Plan framework. The overall compound error along an action chain is significantly reduced without the predictive errors introduced by states.

Getting rid of intermediate state supervision does not mean all sorts of visual information are useless. In procedure planning, the initial visual observation $v_{start}$ and the goal visual observation $v_{goal}$ are vital because action sequences are constrained by them. These visual observations are in the form of frame features in seconds. Inspired by the image analysis above, we argue the visual input information is underutilized by the previous works, especially along the time dimension. In the Skip-Plan framework, we take a shared multi-layer perceptrons (MLP) network in the Visual Input Module to learn the time information, where the MLP network is three-layered with ReLU activations. This simple method can boost the model performance by a large margin, especially for short sequences ($T=3$). 

% \noindent \textbf{Initial and Goal Visual Information.} Getting rid of intermediate state supervision does not mean all sorts of visual information are useless. In procedure planning, the initial visual observation $v_{start}$ and the goal visual observation $v_{goal}$ are vital because action sequences are constrained by them. These visual observations are injected into the network in the form of frame features in seconds. Inspired by the image analysis above, we argue the visual input information is underutilized by the previous works, especially along the time dimension. In the Skip-Plan framework, we take a shared multi-layer perceptrons (MLP) network in the Visual Input Module to learn the time information, where the MLP network is three-layered with ReLU activations. This simple method can boost the model performance by a large margin, especially for short sequences ($T=3$). 

\begin{figure*}[t]
  \centering
   \includegraphics[width=\linewidth]{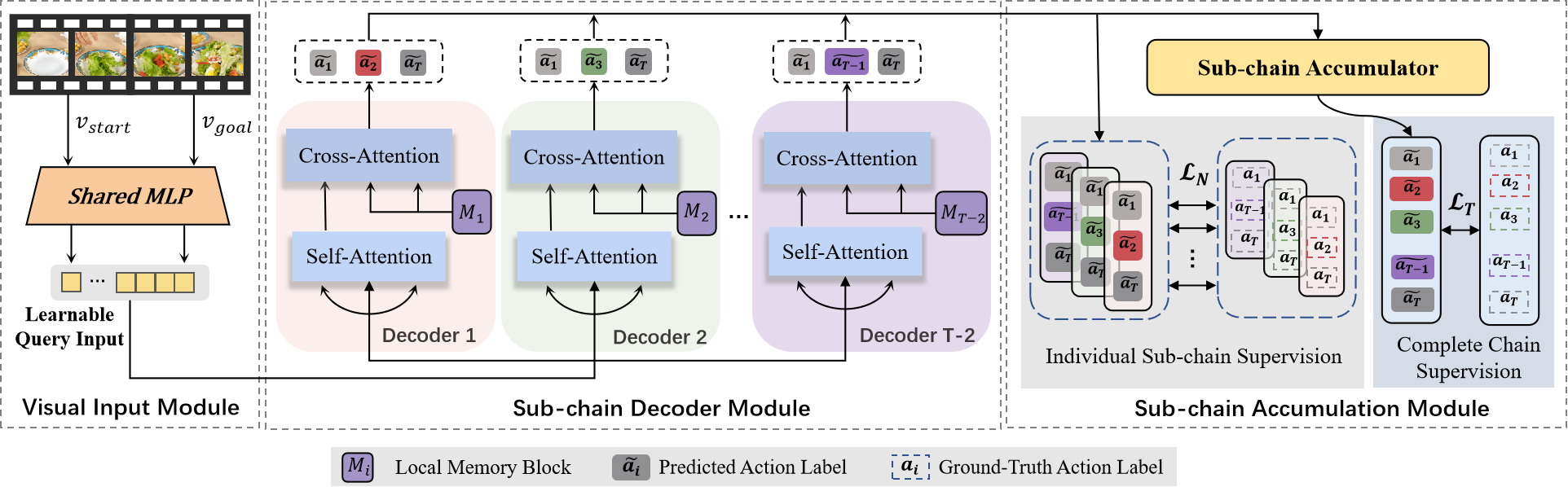}
   % \caption{\textbf{Our Skip-Plan architecture}. The shared MLP network in the Visual Input Module extracts the initial and final information from the initial and goal observations ($v_{start}$ and $v_{goal}$) and attaches them to every learnable query input. A number of $T-2$ query inputs are separately passed to each decoder. The $n$th memory-augmented decoder outputs the sub-chain $\{\Tilde{a_1}, \Tilde{a_t}, \Tilde{a_T}\}$ with $t=n-1$. The predicted sub-chains are supervised by two loss supervision. In Individual Sub-chain Supervision, the predicted sub-chains are supervised by the corresponding ground-truth sub-chains. In Complete Chain Supervision, the predicted sub-chains are aggregated by the Sub-chain Accumulator to generate one complete action chain, which is supervised by the complete ground-truth action chain. }
  \caption{\textbf{Our Skip-Plan architecture}. The shared MLP network in the Visual Input Module extracts the initial and final information from the initial and goal observations ($v_{start}$ and $v_{goal}$) and attaches them to the first and last dimensions of every learnable query input. A number of $T-2$ query inputs are separately passed to each decoder in the Sub-chain Decoder Module. In each decoder, the input is processed by the self-attention block and then by the cross-attention block, where the cross-attention block is augmented by a learnable memory $M_i$. The sub-chain decoders are separated. The $n$th decoder outputs the sub-chain $\{\Tilde{a_1}, \Tilde{a_t}, \Tilde{a_T}\}$ with $t=n+1$. The predicted sub-chains are supervised by two loss supervision. In Individual Sub-chain Supervision, the predicted sub-chains are supervised by the corresponding ground-truth sub-chains. In Complete Chain Supervision, the predicted sub-chains are aggregated by the Sub-chain Accumulator to generate the complete action chain, which is supervised by the complete ground-truth action chain. }
   \label{fig:2}
\end{figure*}

\subsection{Decomposition of Long Chains}
\label{subsec:decom}
Long chains are not favored when a compound error is accumulating. Thus, decomposing long chains into short ones is a natural way to better model performance. Through the lens of mathematics, the only design philosophy here is that we should extract most reliable sub-chains from the original long chain. Following this principle, we ought to find there is so much flexibility in the decoupling method: We can choose sub-chains with any lengths, and actions in the sub-chains are not necessary to be adjacent. 

The reliability of a sub-chain can be determined by the uncertainty of predictions. According to the uncertainty analysis, the reliabilities of the initial and final actions are the highest because they are the closest actions to the initial and goal visual inputs. Since the compound error is accumulated from two ends of the chain, the reliability of action predictions is decreasing while approaching the middle of the chain. Consequently, we remove all the edges between adjacent intermediate actions, and each individual intermediate action only connects to the reliable initial and final actions. This design just coincides with the term $p(a_t|a_1,a_T)$ with $2\leq t \leq T-1$ in Equation \ref{eq:1}. In this way, we generate a quantity of $T-2$ sub-chains from an action chain with a time horizon of $T$. The set of sub-chains is given by:
\begin{equation}
  \{ (a_1, a_2, a_T), (a_1, a_3, a_T), ... , (a_1, a_{T-1}, a_T) \}.
  \label{eq:3}
\end{equation} 
Each sub-chain has a length of 3, with the order of the initial action, one unique intermediate action, and the final action. This decoupling method works well for relatively long chains ($T>3$).

Given multiple short sub-chains, we design a compatible network to model all their sub-relations. Since these sub-chains are parallel and independent to each other, we propose the Sub-chain Decoder Module, where each transformer decoder only models one discriminative relation of one sub-chain. Every transformer decoder is non-autoregressive with two types of inputs: a learnable query with the initial and goal visual features, and a learnable memory component. The novelty here is each decoder only learns one sub-relation, with a unique loss only corresponding to the three actions of that sub-chain. The query input, parameters, and memory component of every decoder are separated from other decoders'. Therefore, the number of decoders in the Sub-chain Decoder Module is equal to the number of sub-chains ($T-2$) extracted from the original chain. The final sequence generation is aggregated from all the sub-chain predictions through the Sub-chain Accumulator, which is an MLP network. In this way, our Skip-Plan network can model all sorts of discriminative sub-relations within the original action sequence.

\subsection{Training}
\label{subsec: training}
Loss is an essential component to discriminate sub-chain decoders. Instead of conventional cross-entropy loss, we use the focal loss to supervise our sub-chain decoders. The focal loss $FL$ is formulated as\cite{focalloss}: 
\begin{equation}
  FL = -\sum_{t=1}^{T} \, a_t \, ( 1 - \Tilde{a}_t )^\gamma \, log \, \Tilde{a}_t,
  \label{eq:4}
\end{equation} 
where we denote $a_t$ as the ground-truth one-hot labels, $\Tilde{a}_t$ as the predicted probability distribution of the action sequence, and $\gamma$ as the gamma value of focal loss.  The focal loss is equivalent to the cross-entropy loss when $\gamma = 0$. The model training is benefited from the use of focal loss because the data distribution in procedure planning is imbalanced. In this work, we define $FL(a_{1}, a_{t}, a_{T})$ as the focal loss for the sub-chain with a length of 3, where the sub-chain is composed of the initial action ${a_1}$, the action at timestep $t$ ${a_t}$, and the final action ${a_T}$. We present the focal loss of the whole action sequence as $FL(a_{1:T})$.  

\noindent \textbf{Loss of Individual Sub-chains.} As we have discussed before, the loss of each decoder only corresponds to the actions within that sub-chain. Therefore, the individual loss of the $n$th decoder is defined as $FL(a_1, a_t, a_T)$ with $t=n+1$, and the loss from all sub-chain decoders can be written as: 
\begin{equation}
  \mathcal{L}_{N} = \sum_{t=2}^{T-1} FL(a_1, a_t, a_T).
  \label{eq:5}
\end{equation} 

\noindent \textbf{Loss of Complete Chain.} After aggregating all sub-chains by the Sub-chain Accumulator, the sequence output with a length of $T$ is supervised by the focal loss of the whole action sequence: 
\begin{equation}
  \mathcal{L}_{T} = FL(a_{1:T}).
  \label{eq:6}
\end{equation}

\noindent \textbf{Overall Loss.} Therefore, the total loss is defined as the sum of the loss from all sub-chains and the complete chain loss, which can be written as:
\begin{equation}
  \mathcal{L} = \mathcal{L}_{N} + \mathcal{L}_{T}.
  \label{eq:7}
\end{equation}
In this way, our model can explore all sorts of discriminative but reliable sub-relations within an action sequence, while guaranteeing the final sequence generation is optimal. 

% Basically, the parameters of the sub-chain decoders are separated. The difference between the decoders is the sub-chain outputs, which result in different parameters after training. We take the individual sub-chain loss to optimize each decoder, and the complete chain loss to optimize the sub-chain accumulator. In this way, our model can explore all sorts of discriminative but reliable sub-relations within an action sequence, while guaranteeing the final sequence generation is optimal. 

\begin{table*}
  \centering
  \caption{Metric results on CrossTask and COIN for short-horizon predictions ($T\in \{3,4\}$). The '\cmark' and '\xmark' in the column \textit{State Supervision} denote with and without state supervision respectively. The best and second best metric results are coloured in \textcolor{red}{red} and \textcolor{blue}{blue} respectively. }
  \label{table:1}
  \resizebox{1.9\columnwidth}{!}{%
  \begin{tabular}{clccccccc}
    \toprule
    \multirow{2}{*}{Horizons} & \multirow{2}{*}{Models} & \multirow{2}{*}{State Supervision} & \multicolumn{3}{c}{CrossTask}& \multicolumn{3}{c}{COIN} \\
    \cmidrule(r){4-6} \cmidrule(r){7-9}
                         &  & &SR & mAcc & mIoU & SR & mAcc & mIoU \\
    \midrule
    \multirow{5}{*}{T=3} & DDN\cite{procedure_planning1} &\cmark  & 12.18  & 31.29   & 47.48   & 13.90 &  20.19     & 64.78   \\
                         & Plate\cite{plate}  &\cmark    & 16.00  & 36.17    & 65.91  & -  & -    & - \\
                         & Ext-GAIL\cite{procedure_planning2}   &\cmark    &21.27  & 49.46    & 61.70 & -  & -    & -  \\
                         & P3IV\cite{p3iv} &\cmark    &\textcolor{blue}{23.34}  & \textcolor{blue}{49.96}  & \textcolor{blue}{73.89}   &\textcolor{blue}{15.40}  & \textcolor{blue}{21.67}  & \textcolor{blue}{76.31}  \\
                         & Skip-Plan   &\xmark  & \textcolor{red}{28.85}   & \textcolor{red}{61.18}   & \textcolor{red}{74.98}    & \textcolor{red}{23.65}   & \textcolor{red}{47.12}   & \textcolor{red}{78.44}  \\
    \midrule
    \multirow{5}{*}{T=4} & DDN\cite{procedure_planning1}   &\cmark    & 5.97  & 27.10    & 48.46   & 11.13  & 17.71    & 68.06  \\
                         & Plate\cite{plate}   &\cmark  & 14.00  & 35.29    & 55.36   & -  & -    & - \\
                         & Ext-GAIL\cite{procedure_planning2}   &\cmark    & \textcolor{red}{16.41}  & 43.05    & 60.93 & -  & -    & -  \\
                         & P3IV\cite{p3iv}  &\cmark    & 13.40  & \textcolor{blue}{44.16}    & \textcolor{blue}{70.01}  & \textcolor{blue}{11.32}  & \textcolor{blue}{18.85}    & \textcolor{blue}{70.53}    \\
                         & Skip-Plan   &\xmark   & \textcolor{blue}{15.56}  & \textcolor{red}{55.64}    & \textcolor{red}{70.30}    & \textcolor{red}{16.04}  & \textcolor{red}{43.19}  & \textcolor{red}{77.07}  \\
    \bottomrule
  \end{tabular}
  }
\end{table*}

\begin{table}
  \centering
  \caption{Metric results for long-horizon predictions ($T\in\{5,6\}$) on CrossTask and COIN.}
  \label{table:2}
  \resizebox{\columnwidth}{!}{%
  \begin{tabular}{ c | c l c c c}
    \toprule
                        Datasets & Horizons & Models & SR & mAcc & mIoU \\
    \midrule
    \multirow{6}{*}{CrossTask} & \multirow{3}{*}{T=5} & DDN\cite{procedure_planning1}     & 3.10  & -   & -   \\
                         & & P3IV\cite{p3iv} & 7.21  & -  & -   \\
                         & & Skip-Plan   & \textbf{8.55}   & \textbf{52.05}   & \textbf{69.09} \\
        \cmidrule(r){2-6}   
        & \multirow{3}{*}{T=6} & DDN\cite{procedure_planning1}   & 1.20  & -    & -  \\
                       &  & P3IV\cite{p3iv}  & 4.40  & -   & -  \\
                        & & Skip-Plan  & \textbf{5.12}  & \textbf{47.34}    & \textbf{66.11} \\
    \midrule

       \multirow{4}{*}{COIN} & \multirow{2}{*}{T=5}  & P3IV\cite{p3iv} & 4.27  & 10.81  & 68.81  \\
                                &  & Skip-Plan   & \textbf{9.90}   & \textbf{38.99}   & \textbf{76.93} \\
                                \cmidrule(r){2-6}   
                & \multirow{2}{*}{T=6}  &  P3IV\cite{p3iv}  & 4.17  & 8.97  & 67.72 \\
                 &   & Skip-Plan  & \textbf{6.61} & \textbf{33.63}    & \textbf{73.68} \\
    \bottomrule
    
  \end{tabular}
  }
\end{table}

\subsection{Implementation Details}
\label{subsec: impledetail}
For each decoder, the query input length is $T+1$, and the memory size is 128. The numbers of decoder layers, decoder heads, and decoder dimensions are set to be 1, 16, and 1024 respectively. All MLP networks are three-layered with ReLU activations. In the Visual Input Module, our shared MLP network on the time dimension of input frame features has the shape $[3 \rightarrow 6 \rightarrow 1]$. All action classifiers in the Sub-chain Decoder Module are the same MLP networks with the shape $[1024 \rightarrow 512 \rightarrow n_a]$, where $n_a$ is the number of action classes. The Sub-chain Accumulator is an MLP network with the shape $[T(T-2) \rightarrow 3T(T-2) \rightarrow T]$. Models are optimized for 500 epochs with stochastic gradient descent. The initial learning rate is 0.02 and decays by 10 at every 50 epochs after 100 epochs. Experiments are conducted on a single GeForce RTX 3090 GPU. We also test our model using the Mindspore \cite{mindspore}. 

\section{Experiments}
\label{sec:experiments}
% In this section, we demonstrate the model performance of our Skip-Plan on two instructional video datasets. We first discuss the details of the datasets, the metrics, and the training scheme on each dataset in Section \ref{subsec: evaluationprotocol}. Then, we compare our metric results with previous state-of-the-art models’, detailed in Section \ref{subsec: mainresult}. Extensive prerequisite and ablation experiments are carried out to validate the statements and demonstrate the effectiveness of each method in Section \ref{sec:methods}. Finally, we show some case studies to prove our model outperforms state-of-the-art methods and illustrate the effects of different components in Skip-Plan.

\subsection{Evaluation Protocols}
\label{subsec: evaluationprotocol}

\noindent \textbf{Datasets.} We evaluate our model on two real-world instructional video datasets: CrossTask\cite{crosstask} and COIN\cite{coin}. CrossTask contains 2763 videos with a duration of 213 hours. It consists of 18 different tasks with 7.4 steps per task. The COIN dataset spans 12 different domains related to our daily life. It contains 11,827 videos with 3.9 steps per video. The COIN dataset is challenging due to the diversity of its tasks and action labels (180 tasks and 779 action labels). We curate a set of plans by the sliding window approach, and adopt the standard 70\%/30\% to create our train/test splits. The division is done on the generated samples rather than the original video. The data pre-processing procedure in our work is identical to the most widely used protocol in the previous procedure planning works (\eg, DDN \cite{procedure_planning1}, Ext-GAIL \cite{procedure_planning2}, and the main protocol in P3IV \cite{p3iv}) for fair comparisons. 

\noindent \textbf{Metrics.} Following the previous works, we used three metrics to comprehensively evaluate model performance. For each action sequence: (a) mean Intersection over Union (mIoU) regards the predicted and ground-truth action sequences as two sets, and measure the overlap between these sets. The mIoU metric only indicates whether desired actions exist in the predicted action set, agnostic to the order of actions. (b) mean Accuracy (mAcc) evaluates the correctness of the predicted action at every timestep, termed as percentages. Thus, mAcc counts the order of actions, and is more strict than mIoU. (c) Success Rate (SR) evaluates the correctness of the whole action sequence. A predicted sequence is successful only if it matches the ground-truth sequence. The success rate is the proportion of successful samples over all samples, which is more strict than mAcc.

\noindent \textbf{Training Scheme on Each Dataset. } We specify an appropriate gamma value on each dataset to maximize model performance. Through experiments, we find the gamma values of $1.5$ and $0$ are the best parameters for the CrossTask and COIN datasets respectively. This is reasonable because the data distribution in the COIN dataset is more balanced due to a large quantity of action labels in diverse tasks. 

\subsection{Evaluating Procedure Planning}
\label{subsec: mainresult}

\noindent \textbf{Short-horizon Prediction.} We compare the model performance of the short-horizon predictions ($T=3/4)$ in Table \ref{table:1}. As illustrated in Table \ref{table:1}, Skip-Plan significantly outperforms the previous state-of-the-art model at all metrics for $T=3$. In the CrossTask dataset, our SR and mAcc results surpass the results of the second best approach (\eg, P3IV\cite{p3iv}) by 6\% and 11\%. Our method also outperforms P3IV\cite{p3iv} by 8\% on SR and 25\% on mAcc in the COIN dataset. Such incredible improvements strongly prove Skip-Plan without state supervision suppresses predictive uncertainty effectively, and our module design on the visual input extracts initial and goal visual information thoroughly. For $T=4$, Skip-Plan achieves large gains of around 5\%, 24\%, and 7\% in SR, mAcc and mIoU respectively in the COIN dataset. For the CrossTask dataset, the mAcc and mIoU of our Skip-Plan model are the best and the SR is the second best. Particularly, our mAcc is higher than P3IV's\cite{p3iv} by 11.48\%. The methods of no state supervision and decoupling long chains both drive Skip-Plan to succeed at $T=4$. 

\noindent \textbf{Long-horizon Prediction.} We further evaluate Skip-Plan on long-horizon predictions ($T=5/6$). In the COIN dataset, our Skip-Plan performs extraordinarily well on the metric of mAcc, with remarkable improvements of 28\% and 25\% for $T=5/6$, compared to the second best model P3IV\cite{p3iv}. As for the metrics of SR and mIoU, Skip-Plan exceeds P3IV\cite{p3iv} by 6\% and 8\% at $T=5$, and 2\% and 6\% at $T=6$ respectively. The outstanding performance we presented above further validates the success of the decoupling approach. Our Skip-Plan also achieved the best results on SR in the CrossTask dataset, surpassing P3IV\cite{p3iv} by 1.34\% and 0.72\% at $T=5/6$. For the long-horizon predictions, we argue mAcc and mIoU are more reliable metrics to measure the effectiveness of a model. SR is interrupted by a large variance, because data samples become insufficient with a large time horizon. 

% \noindent \textbf{Long-horizon Prediction.} We further evaluate Skip-Plan on long-horizon predictions ($T=5/6$). In the COIN dataset, our Skip-Plan performs extraordinarily well on the metric of mAcc, with remarkable improvements of 28\% and 25\% for $T=5/6$, compared to the second best model P3IV\cite{p3iv}. As for the metrics of SR and mIoU, Skip-Plan exceeds P3IV\cite{p3iv} by 6\% and 8\% at $T=5$, and 2\% and 6\% at $T=6$ respectively. The outstanding performance we presented above further validates the success of the decoupling approach. Our Skip-Plan also achieved the best results on SR in the CrossTask dataset, surpassing P3IV\cite{p3iv} by ??1.34\% and 0.72\%?? at $T=5/6$. For the long-horizon predictions, we argue mAcc and mIoU are the more reliable metrics to measure the effectiveness of a model. SR is interrupted by a large variance, because data samples become insufficient with a large time horizon. 

\subsection{Prerequisite Experiments}
\label{subsec: prerequisiteexperiments}
In this section, we carried out several prerequisite experiments for the long-chain decoupling method. They can validate some statements that we have made in Section \ref{subsec:decom}. These experiments are foundational because they proved the prerequisite conditions for the decoupling approach to work. Thus, we name them as the prerequisite experiments. 

\begin{figure}[t]
    \centering
    \includegraphics[width=\linewidth]{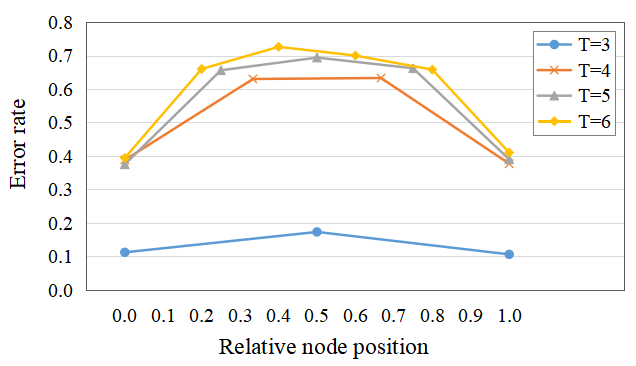}
    \caption{Error rate distributions along the non-autoregressively predicted chains with $T=3/4/5/6$. }
    \label{fig:3}
\end{figure}

\noindent \textbf{Error Rate Distribution along a Predicted Chain. } We tend to remove the edges between unreliable nodes and only connect one unreliable node to other reliable ones. In this way, we can filter out most reliable sub-chains from the original long chain. Here, we measure the reliability of each node in terms of the error rate distribution along the non-autoregressively predicted chain. The error rate distributions along the predicted chains with $T=3/4/5/6$ are illustrated in Figure \ref{fig:3}. It is clear the reliabilities of the first and final action nodes are the highest and the model is more likely to make wrong predictions while approaching the middle of the chain. It is not hard to understand this phenomenon: a higher error rate occurs at intermediate action predictions because they are further away from the initial and goal visual observations. This empirical finding motivates us to remove the edges between unreliable intermediate actions and only connect each intermediate action to the reliable initial and final action nodes. 

\begin{table}
  \centering
  \caption{Reliability of standalone sub-chains vs sub-chains contained within the original long chain.}
  \label{table:3}
  \resizebox{\columnwidth}{!}{%
  \begin{tabular}{ccccc}
    \toprule
                        Sub-chain  & Loss Type & SR & mAcc & mIoU \\
    \midrule
                         \multirow{2}{*}{$\{a_1, a_2, a_5\}$} & Long  & 22.52  & 58.28   & 71.87   \\
                          & Short & \textbf{22.68} & \textbf{58.73}  & \textbf{72.40} \\
    \midrule
                         \multirow{2}{*}{$\{a_1, a_3, a_5\}$} & Long  & 13.73  & 56.74   & 71.78   \\
                          & Short & \textbf{16.29} & \textbf{59.91}  & \textbf{73.07} \\
    \midrule
                         \multirow{2}{*}{$\{a_1, a_4, a_5\}$} & Long  & 12.50  & 56.58   & 71.23   \\
                          & Short & \textbf{13.25} & \textbf{57.61}  & \textbf{71.72} \\
    \bottomrule
  \end{tabular}
  }
\end{table}

\noindent \textbf{Reliability of Standalone Sub-chains. } In addition to the reliability of each node, we further study the reliability of the sub-chains that we have decoupled from the original long chain.  For a clearer explanation, we take a chain with a length of 5 as an example (\eg, $\{a_1, a_2, a_3, a_4, a_5\}$), and only study the reliability of the sub-chain $\{a_1, a_3, a_5\}$. We consider two cases: First, we just model the relation of the whole chain and only investigate the prediction result of that sub-chain contained in the long chain. In other words, we formulate the loss function along the whole sequence (involving $\{a_1, a_2, a_3, a_4, a_5\}$), and only compute the metric results over the sub-chain $\{a_1, a_3, a_5\}$. The second case is we only model the relation of the standalone sub-chain and check its prediction result. Equivalently, we model the loss function just involving $\{a_1, a_3, a_5\}$, and still computes the metric results of the sub-chain $\{a_1, a_3, a_5\}$. The metric results of these two cases are presented in Table \ref{table:3}, where we denote `Long' as the first case and denote `Short' as the second case due to the length difference in loss. We spot the metric results of standalone sub-chains (`Short') are always better than the ones contained within the whole action chain (`Long'). This empirical finding also happens in the cases of $T=4$ and $T=6$. As we stated at the beginning of this section, such experimental results are foundational: The method of decoupling long chains can be only workable when standalone sub-chains can result in better metric results compared to the sub-chains contained within their original long chain. In this way, we experimentally prove the reliability of standalone sub-chains is better under any condition. It constructs the prerequisite condition to guarantee our decoupling method is workable. 

% \subsection{Visualization of Reliability in an Action Chain}
\subsection{Ablation Study}
\label{subsec: ablationstudy}

\begin{table}
  \centering
  \caption{Ablation experiments on CrossTask. }
  \label{table:4}
  \resizebox{\columnwidth}{!}{%   
  \begin{tabular}{l|ccc}
    \toprule
    \makebox[0.1\textwidth][c]{} & \makebox[0.08\textwidth][c]{SR} & \makebox[0.1\textwidth][c]{mAcc}& \makebox[0.06\textwidth][c]{mIoU} \\
                          % & SR & mAcc & mIoU \\
    \midrule
                        \multicolumn{4}{l}{(a) With and Without State Supervision at $T=3$}  \\ 
    \midrule
                         Language   & 26.28  & 58.98   & 71.82   \\
                         None   & \textbf{28.85}   & \textbf{61.18}   & \textbf{74.98} \\
    \midrule
                        \multicolumn{4}{l}{(b) Types of layers on the Time of Input Frames at $T=3$}  \\ 
    \midrule
                         Average Pooling   & 24.40  & 56.41   & 70.87   \\
                         Linear   & 24.40  & 56.69   & 71.24   \\
                         Conv   & 26.56  & 59.60   & 74.15   \\
                         MLP   & \textbf{28.85}   & \textbf{61.18}   & \textbf{74.98} \\
    \midrule
                        \multicolumn{4}{l}{(c) Without and With Decoupling at $T=4$}  \\ 
    \midrule
                         w/o Decoupling   & 12.72  & 53.18   & 69.13  \\
                         Decoupling   & \textbf{15.56}   & \textbf{55.64}   & \textbf{70.30} \\
    \midrule
                        \multicolumn{4}{l}{(d) Without and With Decoupling at $T=5$}  \\ 
    \midrule
                         w/o Decoupling   & 4.97  & 48.71   & 65.29  \\
                         Decoupling   & \textbf{8.55}   & \textbf{52.05}   & \textbf{69.09} \\
    \midrule
                        \multicolumn{4}{l}{(e) Without and With Decoupling at $T=6$}  \\ 
    \midrule
                         w/o Decoupling   & 2.62  & 44.08   & 62.07  \\
                         Decoupling   & \textbf{5.12}   & \textbf{47.34}   & \textbf{66.11} \\
    \midrule
                        \multicolumn{4}{l}{(f) Gamma Values of Focal Loss at $T=4$}  \\ 
    \midrule
                         $\gamma=0$   & 13.84  & 54.67  & 69.74 \\
                         $\gamma=1.0$ & 15.08 & 54.58 & 70.09 \\
                         $\gamma=1.5$   & \textbf{15.56}   & \textbf{55.64}   & \textbf{70.30} \\
                         $\gamma=2.0$ & 14.80 & 54.99 & 70.20 \\
    \bottomrule
  \end{tabular}
  }
\end{table}

We present comprehensive ablation experiments to illustrate the effectiveness of different components in our Skip-Plan. All ablation experiments are carried out on the CrossTask dataset, illustrated in Table \ref{table:4}. 

\noindent \textbf{With and Without State Supervision.} We consider $T=3$ to compare the metric results with and without intermediate state supervision. The model with intermediate state supervision is done by adding weak language state supervision, which is identical to the one proposed by P3IV\cite{p3iv}. The weak language state supervision in P3IV\cite{p3iv} is the most effective state supervision among all the previous works. In Table \ref{table:4}(a), we can see the model without state supervision outperforms the model with language state supervision in all metrics. It confirms the existence of intermediate state nodes does not provide much valuable information but introduces undesired noise into the chain. 

\noindent \textbf{Types of Layers on the Time of Input Frames.} Our model achieves significant improvements by applying a small MLP network on the time dimension of visual input frames. Here, we explore the effects of various types of layers on the time information, including average pooling, a single linear layer, 1D convolution, and our proposed MLP, as shown in Table \ref{table:4}(b). All the previous works overlooked the significance of initial and goal visual inputs. They used the averaging pooling operation, which performs the worst among all types of layers. We take $T=3$ as an example to compare different layers, and indicate the MLP network in our Skip-Plan framework performs the best.  

\noindent \textbf{Decomposition of Long Chains. } We compare the performance without and with the decoupling approach at all long $T$, ranging from 4 to 6. Complete ablation experiments demonstrate the decoupling approach in our Skip-Plan can robustly improve all the metric results at any condition. 

\noindent \textbf{Focal Loss with Gamma Values} We compare the experimental results with different gamma values of focal loss. We find the gamma value of $1.5$ works best to overcome the imbalanced data distribution in the CrossTask dataset. 

% \noindent \textbf{Focal Loss with Gamma Values} We compare the experimental results with different gamma values of focal loss. We find the gamma value of ?$1.0$? works best to overcome the imbalanced data distribution in the CrossTask dataset. 

\subsection{Case Study}
\begin{figure}[t]
    \centering
    \includegraphics[width=1\linewidth]{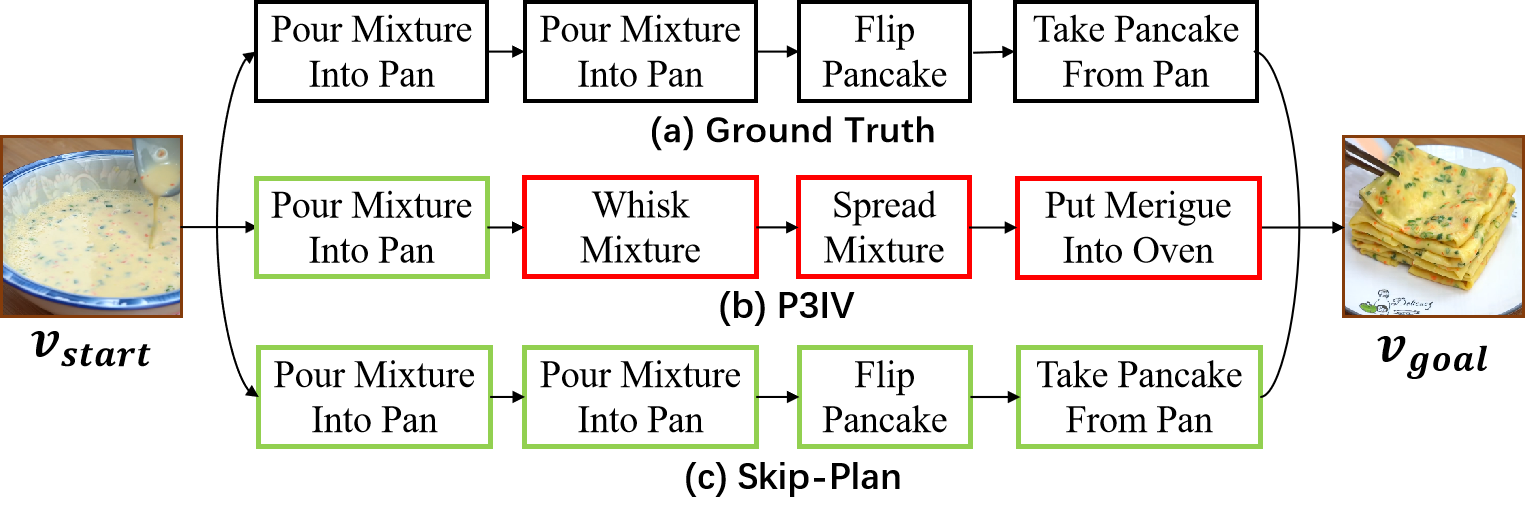}
    \caption{Case study on CrossTask: our Skip-Plan predicts all actions correctly, whereas P3IV\cite{p3iv} fails on the last three predictions. The outlines of correct and incorrect predictions are coloured in green and red respectively.}
    \label{fig:4}
\end{figure}

\noindent \textbf{Skip-Plan and the Baseline. }We compare our Skip-Plan with the previous state-of-the-art P3IV\cite{p3iv} in real cases, illustrated in Figure \ref{fig:4}. Given the initial and goal visual observations, P3IV\cite{p3iv} only predicts the first action correctly and fails to get the other actions at $T=4$. In contrast, Skip-Plan extracts the initial and goal visual information thoroughly, so Skip-Plan predicts the final action correctly. Benefiting from the decoupling method, our Skip-Plan makes the correct predictions on the second and third actions. 

% \section{Case Study on Different Components}
% \label{sec: casecomponent}

\begin{figure}[t]
    \centering
    \includegraphics[width=\linewidth]{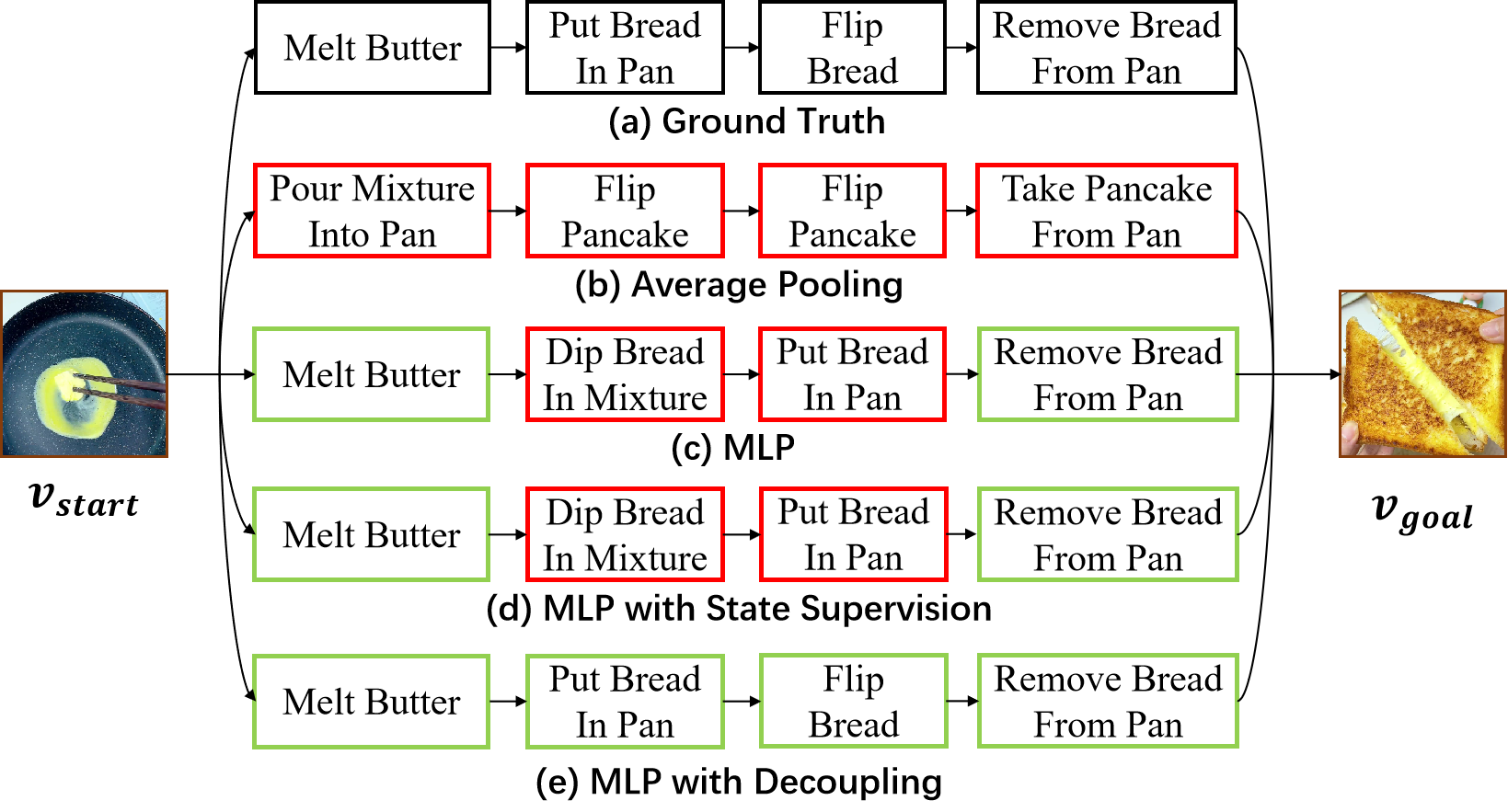}
    \caption{Case study on Skip-Plan components at $T=4$: (a) ground truth; (b) average pooling: average pooling layer on the time dimension of visual input frames without decoupling and state supervision; (c) MLP: shared MLP network on the time dimension of visual input frames without decoupling and state supervision; (d) MLP with state supervision: shared MLP network with state supervision but without decoupling; (e) MLP with decoupling: shared MLP network with decoupling but without state supervision. The outlines of correct and incorrect action predictions are coloured in green and red respectively. }
    \vspace{-1pt}
    \label{fig:S3}
\end{figure}

\noindent \textbf{Effects of Different Components in Skip-Plan. }We investigate the effects of various components in Skip-Plan on model performance by real samples, which are illustrated in Figure \ref{fig:S3}. Here, the components for comparisons include the types of layers on the time dimension of visual input frames, state supervision, and the decoupling approach. For the types of layers on the time dimension of visual input frames, we compare the average pooling layer with our shared MLP network, both without state supervision and decoupling. As shown in Figure \ref{fig:S3}(b), the average pooling layer makes wrong predictions on all actions. However, the model with the shared MLP network predicts the first and last actions correctly, as illustrated in Figure \ref{fig:S3}(c). This is because the MLP network can extract the time information from the initial and goal visual observations more thoroughly than the average pooling layer. We further simulate the shared MLP network without decoupling but with language state supervision\cite{p3iv} in Figure \ref{fig:S3}(d). Comparing Figure \ref{fig:S3}(d) with Figure \ref{fig:S3}(c), we notice the addition of state supervision cannot correct the predictions of the second and the third actions, because the state supervision does not reduce the compound error. In contrast, the shared MLP network without state supervision but with decoupling can lower the compound error effectively, resulting in correct predictions of all actions, as demonstrated in Figure \ref{fig:S3}(e). 

\section{Conclusion}
\label{sec:conclusion}
In this paper, we propose Skip-Plan, a condensed action space learning method for the procedure planning task. Our Skip-Plan solves the procedure planning problem mathematically in two novel ways. First, Skip-Plan skips the intermediate state supervision and only focuses on action relations. Second, Skip-Plan decomposes long action chains into multiple reliable sub-chains by skipping most uncertain intermediate actions. In these ways, our Skip-Plan model explores all sorts of sub-relations within an action chain and generates actions non-adjacently only in the condensed action space. Extensive experiments show the superior performance of our Skip-Plan method on two instructional video datasets. Skip-Plan has the potential to inspire the research field of procedure planning, since most of them heavily rely on state observations and adjacent actions at all times. 

% \textbf{Acknowledgement. }This work was supported in part by National Natural Science Foundation of China (Grant No. 62206153), CAAI-Huawei MindSpore Open Fund. 

\textbf{Acknowledgement. }This work was supported in part by the National Key Research and Development Program of China under Grant 2022ZD0114903 and in part by the National Natural Science Foundation of China under Grant 62125603 and Grant 62206153, CAAI-Huawei MindSpore Open Fund. 

% In this paper, we propose Skip-Plan, a condensed action space learning method for the procedure planning task. Our Skip-Plan solves the procedure planning problem mathematically in two novel ways. First, our Skip-Plan skips the intermediate state supervision and only focuses on action relations. Second, our Skip-Plan decomposes long action chains into multiple reliable sub-chains by skipping most uncertain intermediate actions. In these ways, our Skip-Plan model explores all sorts of sub-relations within an action chain and generates actions non-adjacently only in the condensed action space. Extensive experiments show the superior performance of our method on two instructional video datasets. Our Skip-Plan framework has the potential to inspire any type of action-planning research field, since most of them heavily rely on state observations and adjacent actions at all times. 

{\small
\bibliographystyle{ieee_fullname}
\bibliography{egbib}

\begin{thebibliography}{10}\itemsep=-1pt

\bibitem{mindspore}
Mindspore.
\newblock \url{. https://www.mindspore.cn/}.

\bibitem{uncertainty6}
Arsenii Ashukha, Alexander Lyzhov, Dmitry Molchanov, and Dmitry Vetrov.
\newblock Pitfalls of in-domain uncertainty estimation and ensembling in deep learning.
\newblock {\em arXiv preprint arXiv:2002.06470}, 2020.

\bibitem{procedureseg3}
Anil Batra, Shreyank Gowda, Laura Sevilla-Lara, and Frank Keller.
\newblock A closer look at temporal ordering in the segmentation of instructional videos.
\newblock {\em arXiv preprint arXiv:2209.15501}, 2022.

\bibitem{procedure_planning2}
Jing Bi, Jiebo Luo, and Chenliang Xu.
\newblock Procedure planning in instructional videos via contextual modeling and model-based policy learning.
\newblock In {\em ICCV}, pages 15611--15620, 2021.

\bibitem{procedureseg4}
Nicolas Biard, Salom{\'e} Cojean, and Eric Jamet.
\newblock Effects of segmentation and pacing on procedural learning by video.
\newblock {\em Comput. Hum. Behav.}, 89:411--417, 2018.

\bibitem{detr}
Nicolas Carion, Francisco Massa, Gabriel Synnaeve, Nicolas Usunier, Alexander Kirillov, and Sergey Zagoruyko.
\newblock End-to-end object detection with transformers.
\newblock In {\em ECCV}, pages 213--229, 2020.

\bibitem{procedure_planning1}
Chien-Yi Chang, De-An Huang, Danfei Xu, Ehsan Adeli, Li Fei-Fei, and Juan~Carlos Niebles.
\newblock Procedure planning in instructional videos.
\newblock In {\em ECCV}, pages 334--350, 2020.

\bibitem{videosum1}
Chekuri Choudary and Tiecheng Liu.
\newblock Summarization of visual content in instructional videos.
\newblock {\em IEEE Trans. Multimed.}, 9(7):1443--1455, 2007.

\bibitem{bert}
Jacob Devlin, Ming-Wei Chang, Kenton Lee, and Kristina Toutanova.
\newblock Bert: Pre-training of deep bidirectional transformers for language understanding.
\newblock {\em arXiv preprint arXiv:1810.04805}, 2018.

\bibitem{uncertainty7}
Terrance DeVries and Graham~W Taylor.
\newblock Improved regularization of convolutional neural networks with cutout.
\newblock {\em arXiv preprint arXiv:1708.04552}, 2017.

\bibitem{steploc2}
Mikita Dvornik, Isma Hadji, Konstantinos~G Derpanis, Animesh Garg, and Allan Jepson.
\newblock Drop-dtw: Aligning common signal between sequences while dropping outliers.
\newblock {\em NeuRIPS}, 34:13782--13793, 2021.

\bibitem{uncertaintysurvey}
Jakob Gawlikowski, Cedrique Rovile~Njieutcheu Tassi, Mohsin Ali, Jongseok Lee, Matthias Humt, Jianxiang Feng, Anna Kruspe, Rudolph Triebel, Peter Jung, Ribana Roscher, et~al.
\newblock A survey of uncertainty in deep neural networks.
\newblock {\em arXiv preprint arXiv:2107.03342}, 2021.

\bibitem{actionseg3}
Reza Ghoddoosian, Isht Dwivedi, Nakul Agarwal, Chiho Choi, and Behzad Dariush.
\newblock Weakly-supervised online action segmentation in multi-view instructional videos.
\newblock In {\em CVPR}, pages 13780--13790, 2022.

\bibitem{actionseg4}
Reza Ghoddoosian, Saif Sayed, and Vassilis Athitsos.
\newblock Hierarchical modeling for task recognition and action segmentation in weakly-labeled instructional videos.
\newblock In {\em WACV}, pages 1922--1932, 2022.

\bibitem{uncertainty8}
Dan Hendrycks and Kevin Gimpel.
\newblock A baseline for detecting misclassified and out-of-distribution examples in neural networks.
\newblock {\em arXiv preprint arXiv:1610.02136}, 2016.

\bibitem{videosum5}
Gabriel Huang, Bo Pang, Zhenhai Zhu, Clara Rivera, and Radu Soricut.
\newblock Multimodal pretraining for dense video captioning.
\newblock {\em arXiv preprint arXiv:2011.11760}, 2020.

\bibitem{uncertainty3}
Eyke H{\"u}llermeier and Willem Waegeman.
\newblock Aleatoric and epistemic uncertainty in machine learning: An introduction to concepts and methods.
\newblock {\em Mach. Learn}, 2021.

\bibitem{videosum6}
Vladimir Iashin and Esa Rahtu.
\newblock Multi-modal dense video captioning.
\newblock In {\em CVPRW}, pages 958--959, 2020.

\bibitem{procedureseg2}
Lei Ji, Chenfei Wu, Daisy Zhou, Kun Yan, Edward Cui, Xilin Chen, and Nan Duan.
\newblock Learning temporal video procedure segmentation from an automatically collected large dataset.
\newblock In {\em WACV}, pages 1506--1515, 2022.

\bibitem{uncertainty5}
Alex Kendall and Yarin Gal.
\newblock What uncertainties do we need in bayesian deep learning for computer vision?
\newblock {\em NeurIPS}, 2017.

\bibitem{actionseg1}
Colin Lea, Michael~D Flynn, Rene Vidal, Austin Reiter, and Gregory~D Hager.
\newblock Temporal convolutional networks for action segmentation and detection.
\newblock In {\em CVPR}, pages 156--165, 2017.

\bibitem{actionseg2}
Muheng Li, Lei Chen, Yueqi Duan, Zhilan Hu, Jianjiang Feng, Jie Zhou, and Jiwen Lu.
\newblock Bridge-prompt: Towards ordinal action understanding in instructional videos.
\newblock In {\em CVPR}, pages 19880--19889, 2022.

\bibitem{uncertainty9}
Shiyu Liang, Yixuan Li, and Rayadurgam Srikant.
\newblock Enhancing the reliability of out-of-distribution image detection in neural networks.
\newblock {\em arXiv preprint arXiv:1706.02690}, 2017.

\bibitem{focalloss}
Tsung-Yi Lin, Priya Goyal, Ross Girshick, Kaiming He, and Piotr Doll{\'a}r.
\newblock Focal loss for dense object detection.
\newblock In {\em ICCV}, pages 2980--2988, 2017.

\bibitem{swin}
Ze Liu, Yutong Lin, Yue Cao, Han Hu, Yixuan Wei, Zheng Zhang, Stephen Lin, and Baining Guo.
\newblock Swin transformer: Hierarchical vision transformer using shifted windows.
\newblock In {\em ICCV}, pages 10012--10022, 2021.

\bibitem{steploc5}
Huaishao Luo, Lei Ji, Botian Shi, Haoyang Huang, Nan Duan, Tianrui Li, Jason Li, Taroon Bharti, and Ming Zhou.
\newblock Univl: A unified video and language pre-training model for multimodal understanding and generation.
\newblock {\em arXiv preprint arXiv:2002.06353}, 2020.

\bibitem{uncertainty2}
Andrey Malinin and Mark Gales.
\newblock Predictive uncertainty estimation via prior networks.
\newblock {\em NeurIPS}, 2018.

\bibitem{actionseg5}
Antoine Miech, Jean-Baptiste Alayrac, Lucas Smaira, Ivan Laptev, Josef Sivic, and Andrew Zisserman.
\newblock End-to-end learning of visual representations from uncurated instructional videos.
\newblock In {\em CVPR}, pages 9879--9889, 2020.

\bibitem{videosum3}
Medhini Narasimhan, Arsha Nagrani, Chen Sun, Michael Rubinstein, Trevor Darrell, Anna Rohrbach, and Cordelia Schmid.
\newblock Tl; dw? summarizing instructional videos with task relevance and cross-modal saliency.
\newblock In {\em ECCV}, pages 540--557, 2022.

\bibitem{uncertainty4}
Yaniv Ovadia, Emily Fertig, Jie Ren, Zachary Nado, David Sculley, Sebastian Nowozin, Joshua Dillon, Balaji Lakshminarayanan, and Jasper Snoek.
\newblock Can you trust your model's uncertainty? evaluating predictive uncertainty under dataset shift.
\newblock {\em NeurIPS}, 32, 2019.

\bibitem{actionseg6}
AJ Piergiovanni, Anelia Angelova, Michael~S Ryoo, and Irfan Essa.
\newblock Unsupervised action segmentation for instructional videos.
\newblock {\em arXiv preprint arXiv:2106.03738}, 2021.

\bibitem{uncertainty10}
Alireza Shafaei, Mark Schmidt, and James~J Little.
\newblock A less biased evaluation of out-of-distribution sample detectors.
\newblock {\em arXiv preprint arXiv:1809.04729}, 2018.

\bibitem{steploc4}
Anshul Shah, Benjamin Lundell, Harpreet Sawhney, and Rama Chellappa.
\newblock Steps: Self-supervised key step extraction from unlabeled procedural videos.
\newblock {\em arXiv preprint arXiv:2301.00794}, 2023.

\bibitem{videosum4}
Botian Shi, Lei Ji, Yaobo Liang, Nan Duan, Peng Chen, Zhendong Niu, and Ming Zhou.
\newblock Dense procedure captioning in narrated instructional videos.
\newblock In {\em ACL}, pages 6382--6391, 2019.

\bibitem{videosum2}
Chen Sun, Austin Myers, Carl Vondrick, Kevin Murphy, and Cordelia Schmid.
\newblock Videobert: A joint model for video and language representation learning.
\newblock In {\em ICCV}, pages 7464--7473, 2019.

\bibitem{plate}
Jiankai Sun, De-An Huang, Bo Lu, Yun-Hui Liu, Bolei Zhou, and Animesh Garg.
\newblock Plate: Visually-grounded planning with transformers in procedural tasks.
\newblock {\em IEEE Robot. Autom. Lett.}, 7(2):4924--4930, 2022.

\bibitem{coin}
Yansong Tang, Dajun Ding, Yongming Rao, Yu Zheng, Danyang Zhang, Lili Zhao, Jiwen Lu, and Jie Zhou.
\newblock Coin: A large-scale dataset for comprehensive instructional video analysis.
\newblock In {\em CVPR}, pages 1207--1216, 2019.

\bibitem{transformer}
Ashish Vaswani, Noam Shazeer, Niki Parmar, Jakob Uszkoreit, Llion Jones, Aidan~N Gomez, {\L}ukasz Kaiser, and Illia Polosukhin.
\newblock Attention is all you need.
\newblock {\em NeurIPS}, 2017.

\bibitem{steploc1}
Mengmeng Xu, Juan-Manuel P{\'e}rez-R{\'u}a, Victor Escorcia, Brais Martinez, Xiatian Zhu, Li Zhang, Bernard Ghanem, and Tao Xiang.
\newblock Boundary-sensitive pre-training for temporal localization in videos.
\newblock In {\em ICCV}, pages 7220--7230, 2021.

\bibitem{tqn}
Chuhan Zhang, Ankush Gupta, and Andrew Zisserman.
\newblock Temporal query networks for fine-grained video understanding.
\newblock In {\em CVPR}, pages 4486--4496, 2021.

\bibitem{p3iv}
He Zhao, Isma Hadji, Nikita Dvornik, Konstantinos~G Derpanis, Richard~P Wildes, and Allan~D Jepson.
\newblock P3iv: Probabilistic procedure planning from instructional videos with weak supervision.
\newblock In {\em CVPR}, pages 2938--2948, 2022.

\bibitem{procedureseg1}
Luowei Zhou, Chenliang Xu, and Jason Corso.
\newblock Towards automatic learning of procedures from web instructional videos.
\newblock In {\em AAAI}, pages 7590--7598, 2018.

\bibitem{steploc3}
Linchao Zhu and Yi Yang.
\newblock Actbert: Learning global-local video-text representations.
\newblock In {\em CVPR}, pages 8746--8755, 2020.

\bibitem{crosstask}
Dimitri Zhukov, Jean-Baptiste Alayrac, Ramazan~Gokberk Cinbis, David Fouhey, Ivan Laptev, and Josef Sivic.
\newblock Cross-task weakly supervised learning from instructional videos.
\newblock In {\em CVPR}, pages 3537--3545, 2019.

\end{thebibliography}
}

\newpage
\section*{Supplemental}
\appendix
We provide additional information and experimental results in the supplemental material. In Section \ref{sec: errorrate}, we present the analysis of the error rate distributions on the action chains predicted by different works. It illustrates the behaviour of error accumulation with concrete experimental evidence. Then, we detail the network architecture and loss design at $T=3$ in Section \ref{sec:St3}. We further demonstrate the robustness and delicacy of our decoupling approach through experiments in Section \ref{sec: reliability}.

\section{Comparisons of Error Rate Distributions}
\label{sec: errorrate}

\begin{figure}[t]
    \centering
    \includegraphics[width=0.98\linewidth]{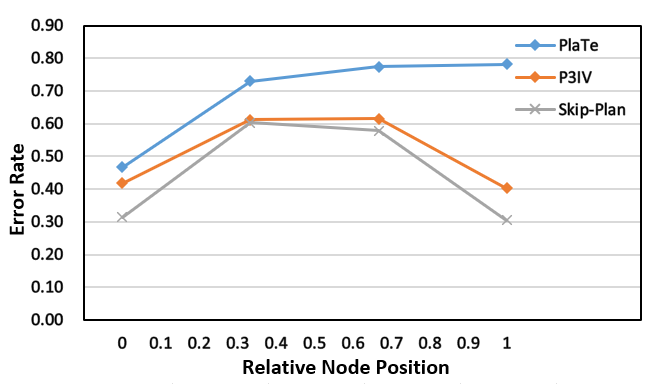}
    \caption{Error rate distributions along the chains predicted by PlaTe\cite{plate}, P3IV\cite{p3iv}, and Skip-Plan at $T=4$. }
    \vspace{-1pt}
    \label{fig:S2}
\end{figure}

Error rate analysis is crucial because it reveals how errors are accumulated along action chains. In this section, we compare the error rate distributions along the chains at $T=4$ predicted by different works, including PlaTe\cite{plate}, P3IV\cite{p3iv}, and our Skip-Plan, as illustrated in Figure \ref{fig:S2}. Here, the error rate at a timestep $t$ is defined as the number of wrong action predictions at the timestep $t$ divided by the total number of actions at this position, and the relative node position is calculated by $(t-1)/(T-1)$. The error rate distribution is the distribution of the error rates at all timesteps (\eg, the relative node position ranges from 0 to 1). We can summarize two important points from Figure \ref{fig:S2}. First, we spot the error rate distribution of PlaTe\cite{plate} is significantly different from the ones of P3IV\cite{p3iv} and Skip-Plan. The error rate distribution of PlaTe\cite{plate} keeps rising until the last action of the chain, but the distributions of P3IV\cite{p3iv} and our Skip-Plan are increasing and then decreasing, reaching a maximum at the middle of the chains. This discrepancy is caused by the network type. PlaTe\cite{plate} is an autoregressive network and generates actions one by one. Thus, it accumulates errors continuously from left to right. In contrast to PlaTe\cite{plate}, P3IV\cite{p3iv} and our Skip-Plan utilize non-autoregressive transformer decoders to generate whole action sequences in batches. This type of network accumulates the error from two ends of the chain, and the error rate peaks at the middle of the chain. Overall, the average error rate of the non-autoregressive models is lower than the one of the autoregressive models, because the chain length for the error accumulation in the non-autoregressive models is reduced to half compared to the length of the autoregressive models. Second, benefiting from the shared MLP network in the Visual Input Module, the error rates of the first and last actions in our Skip-Plan are the lowest among these works. The decoupling approach without state supervision in Skip-Plan further reduces the compounding error at intermediate actions. Both of them drive our Skip-Plan network to achieve the lowest error rates at all timesteps. Consequently, our Skip-Plan achieves state-of-the-art performance on the CrossTask and COIN benchmarks in procedure planning. 

\begin{figure*}[t]
  \centering
  \includegraphics[width=0.8\linewidth]{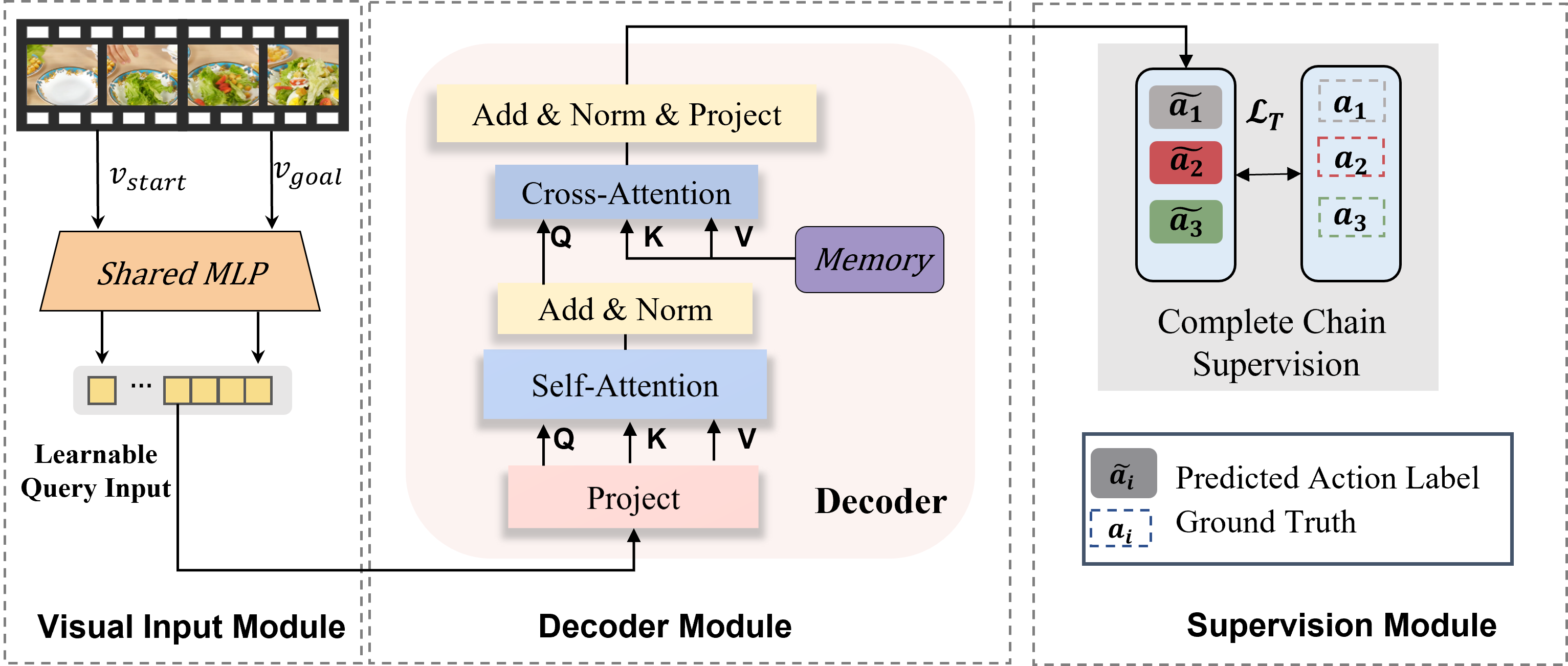}
  \caption{\textbf{Our Skip-Plan architecture at T=3.} The network architecture at $T=3$ is slightly different from Figure 2. It is composed of the Visual Input Module, the Decoder Module, and the Supervision Module, where the Decoder Module and the Supervision Module correspond to the Sub-chain Decoder Module and the Sub-chain Accumulation Module in Figure 2. There is only one decoder in the Decoder Module, which directly outputs the complete chain $\{a_1, a_2, a_3\}$. Without generating sub-chains, there is only one complete chain supervision in the Supervision Module. }
  \vspace{-1pt}
  \label{fig:S1}
\end{figure*}

\section{Network Architecture and Loss for T=3}
\label{sec:St3}

Our Skip-Plan architecture at $T=3$ is slightly different from Figure 2. For $T=3$, the network architecture is composed of the Visual Input Module, the Decoder Module, and the Supervision Module, where the Decoder Module and the Supervision Module just correspond to the Sub-chain Decoder Module and the Sub-chain Accumulation Module in Figure 2 respectively. As illustrated in Figure \ref{fig:S1}, there is only one decoder in the Decoder Module, which directly outputs the complete chain $\{a_1, a_2, a_3\}$. Without generating sub-chains, no individual sub-chain supervision and sub-chain accumulator are present in the Supervision Module. Therefore, we only have the complete chain supervision, and the loss for $T=3$ is defined as:
\begin{equation}
  \mathcal{L} = FL(a_{1:T}).
  \label{eq:S1}
\end{equation}

\section{Reliability of Standalone Sub-chains}
\label{sec: reliability}
\begin{table}
  \centering
  \caption{Reliability of standalone sub-chains vs sub-chains contained within original long chains at $T=4/6$. It validates our decoupling approach can robustly improve the metric results for all long $T$. }
  \label{table:S1}
  \setlength\tabcolsep{4pt}
  \resizebox{\columnwidth}{!}{%
  \begin{tabular}{c|ccccc}
    \toprule
                        Horizon & Sub-chain  & Loss Type & SR & mAcc & mIoU \\
    \midrule
                         \multirow{4}{*}{$T=4$} & \multirow{2}{*}{$\{a_1, a_2, a_4\}$} & Long  & 15.77  & 57.58   & 71.05   \\
                         & & Short & \textbf{17.13} & \textbf{59.16}  & \textbf{72.18} \\
    \cmidrule(r){2-6}
                         & \multirow{2}{*}{$\{a_1, a_3, a_4\}$} & Long  & 15.39  & 57.34   & 72.29   \\
                         & & Short & \textbf{16.88} & \textbf{58.93}  & \textbf{73.06} \\
    \midrule
                         \multirow{8}{*}{$T=6$} & \multirow{2}{*}{$\{a_1, a_2, a_6\}$} & Long  & 20.87  & 55.73   & 69.71   \\
                         & & Short & \textbf{23.34} & \textbf{58.85}  & \textbf{71.87} \\
    \cmidrule(r){2-6}
                         & \multirow{2}{*}{$\{a_1, a_3, a_6\}$} & Long  & 15.83  & 52.76   & 67.38   \\
                         & & Short & \textbf{19.81} & \textbf{56.57}  & \textbf{70.55} \\
    \cmidrule(r){2-6}
                         & \multirow{2}{*}{$\{a_1, a_4, a_6\}$} & Long  & 18.75  & 54.89   & 69.10   \\
                         & & Short & \textbf{18.95} & \textbf{55.21}  & \textbf{70.06} \\
    \cmidrule(r){2-6}
                         & \multirow{2}{*}{$\{a_1, a_5, a_6\}$} & Long  & 21.07  & 55.96   & 70.82   \\
                         & & Short & \textbf{22.48} & \textbf{57.81}  & \textbf{72.12} \\
    \bottomrule
  \end{tabular}
  }
\end{table}

\begin{table}
  \centering
  \caption{To demonstrate the delicacy of our decoupling method, we try a different decoupling approach and show how it fails. We compare the reliability of the standalone sub-chain $\{a_1, a_2, a_3\}$ vs the sub-chain $\{a_1,a_2,a_3\}$ contained within the original long chain at $T=4/5/6$. Without the reliable constraint of the last action, the metric results of 'Short' are worse than the 'Long' results at all long T.}
  \label{table:S2}
  \begin{tabular}{ccccc}
    \toprule
                        Horizon  & Loss Type & SR & mAcc & mIoU \\
    \midrule
                         \multirow{2}{*}{$T=4$} & Long  & \textbf{16.65}  & \textbf{49.30}   & \textbf{67.16}   \\
                          & Short & 15.91 & 48.20  & 65.93 \\
    \midrule
                         \multirow{2}{*}{$T=5$} & Long  & \textbf{13.10}  & \textbf{45.67}   & \textbf{64.46}   \\
                          & Short & 12.82 & 45.13  & 62.78 \\
    \midrule
                         \multirow{2}{*}{$T=6$} & Long  & \textbf{11.64}  & \textbf{44.32}   & \textbf{62.29}   \\
                          & Short & 11.44 & 43.30  & 61.50 \\
    \bottomrule
  \end{tabular}
\end{table}

% \begin{table}
%   \centering
%   \caption{To demonstrate the delicacy of our decoupling method, we try a different decoupling approach and show how it fails. We compare the reliability of the standalone sub-chain $\{a_1, a_2, a_3\}$ vs the sub-chain $\{a_1,a_2,a_3\}$ contained within the original long chain at $T=4/5/6$. Without the reliable constraint of the last action, the metric results of 'Short' are worse than the 'Long' results at all long T.}
%   \label{table:S2}
%   \resizebox{\columnwidth}{!}{%
%   \begin{tabular}{ccccc}
%     \toprule
%                         Horizon  & Loss Type & SR & mAcc & mIoU \\
%     \midrule
%                          \multirow{2}{*}{$T=4$} & Long  & \textbf{16.65}  & \textbf{49.30}   & \textbf{67.16}   \\
%                           & Short & 15.91 & 48.20  & 65.93 \\
%     \midrule
%                          \multirow{2}{*}{$T=5$} & Long  & \textbf{13.10}  & \textbf{45.67}   & \textbf{64.46}   \\
%                           & Short & 12.82 & 45.13  & 62.78 \\
%     \midrule
%                          \multirow{2}{*}{$T=6$} & Long  & \textbf{11.64}  & \textbf{44.32}   & \textbf{62.29}   \\
%                           & Short & 11.44 & 43.30  & 61.50 \\
%     \bottomrule
%   \end{tabular}
%   }
% \end{table}

To prove our decoupling approach can extract reliable sub-chains at any condition, we further compare the reliability of standalone sub-chains with the reliability of these sub-chains contained within the complete chain at $T=4/6$. As illustrated in Table \ref{table:S1}, the metric results of `Short' are consistently better than the `Long' results for all sub-chains at any length. Consequently, our decoupling method can robustly better the prediction results for all long $T$. 

To demonstrate the delicacy of our decoupling method, we try a different decoupling strategy and show how it fails. For example, we choose the sub-chain $\{a_1, a_2, a_3\}$ at $T=4/5/6$. This sub-chain has the same length as our decoupled sub-chains, but is composed of one reliable initial action and two unreliable intermediate actions, where the actions are all adjacent. In this way, we find the reliability of standalone sub-chains is lower than the one of the sub-chains contained within the original long chains at all long $T$, illustrated in Table \ref{table:S2}. Thus, this type of decoupling approach cannot improve prediction accuracy. The failure of this decoupling method is caused by losing the reliable constraint of the last action. This simple experiment demonstrates our decoupling design is very delicate, robust, and effective. 

\end{document}